\documentclass[conference]{IEEEtran}
\IEEEoverridecommandlockouts

\usepackage{color}

\usepackage{multirow}
\usepackage{amsmath}

\usepackage{cite}
\usepackage{amsmath,amssymb,amsfonts}
\usepackage{algorithmic}
\usepackage{graphicx}
\usepackage{textcomp}
\usepackage{xcolor}
\def\BibTeX{{\rm B\kern-.05em{\sc i\kern-.025em b}\kern-.08em
    T\kern-.1667em\lower.7ex\hbox{E}\kern-.125emX}}
\begin{document}

\title{Exploiting Method Names to Improve Code Summarization: A Deliberation Multi-Task Learning Approach
	\thanks{$\dagger$ Corresponding author.}
}

\author{
	Rui Xie$^{1,2}$, Wei Ye$^{1 \dagger}$, Jinan Sun$^{1}$, and Shikun Zhang$^{1}$ \\
	{
		$^1$ National Engineering Research Center for Software Engineering, Peking University, Beijing, China
	} \\
	{
		$^2$School of Software and Microelectronics, Peking University, Beijing, China
	} \\
	\textit{{ruixie, wye, sjn, zhangsk}@pku.edu.cn} \\
}

\maketitle

\begin{abstract}
Code summaries are brief natural language descriptions of source code pieces.  The main purpose of code summarization is to assist developers in understanding code and to reduce documentation workload. In this paper, we design a novel multi-task learning (MTL) approach for code summarization through mining the relationship between method code summaries and method names. More specifically, since a method's name can be considered as a shorter version of its code summary, we first introduce the tasks of generation and informativeness prediction of method names as two auxiliary training objectives for code summarization. 
A novel two-pass deliberation mechanism is then incorporated into our MTL architecture to generate more consistent intermediate states fed into a summary decoder, especially when informative method names do not exist. To evaluate our deliberation MTL approach, we carried out a large-scale experiment on two existing datasets for Java and Python. The experiment results show that our technique can be easily applied to many state-of-the-art neural models for code summarization and improve their performance. Meanwhile, our approach shows significant superiority when generating summaries for methods with non-informative names.
\end{abstract}


\section{Introduction}\label{Introduction}

\begin{figure*}[!t]  
\centering
\includegraphics[width=1.\linewidth]{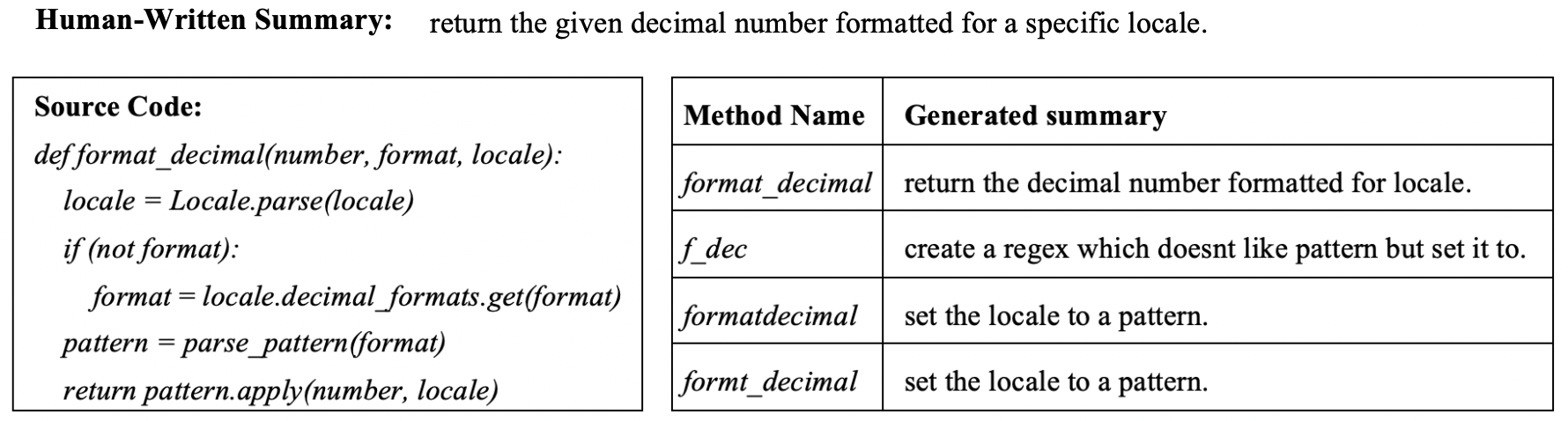}
\caption{Generated Summaries generated by a state-of-the-art model Ast-attendgru~\cite{Leclair2019A} with different method names. The model generates a high-quality summary when we feed an expected informative method name ( \textit{format\_decimal}) into the model. We then replace the method name with ones which involve customized abbreviations(\textit{f\_dec}), nonstandard conjunction(\textit{formatdecimal}), and a spelling error(\textit{formt\_decimal}) to simulate the scenario of summarizing the poorly-named methods. The quality of generated summaries is severely degraded.}
\label{fig:cos_with_diff_name}
\end{figure*}

Source code summarization is the task of creating readable summaries that describe the functionality of source code pieces ~\cite{Mcburney2016Automatic}. High-quality code summarization not only greatly frees programmers from their tedious work on code documentation~\cite{Hu2018Deep}, but also benefits source code maintenance and improves the performance of code retrieval~\cite{Yao2018Improving}. 
The task of code summarization mainly involves two different but related scenarios. The first scenario is to predict a natural language sentence that describes a given method. Researchers have built corresponding datasets by extracting source code of individual methods and the first sentence of their documentation (e.g., docstring of Python method\cite{Yao2018Improving, Barone2017A} and Javadoc of Java method~\cite{Mcburney2016Automatic, Hu2018Deep, Leclair2019A, Hu2018Summarizing}). The second scenario is to predict a natural language sentence that describes a given code snippet~\cite{Alon2018code2seq, Yao2019CoaCor}, e.g., from Stack Overflow, which is not necessarily the source code of an entire method. 
We focus on the first scenario in this paper.


Deep learning is now widely used in code summarization as a mainstream approach. More and more techniques of neural network for machine translation and text summarization have been introduced into code summarization recently ever since Iyer et al.~\cite{Iyer2016Summarizing} proposed a classical neural attention model. Researchers in the fields of natural language processing, deep learning, and software engineering have proposed a variety of neural models, which mainly involve advances in two following directions. The first direction is to design better feature representation to capture more accurate semantics of code such as control-flow structure information of a method~\cite{Hu2018Deep, Alon2018code2seq}. 
The second direction is to bring in more sophisticated deep learning techniques of text generation, such as variational auto encoder~\cite{Chen2018VAE}, dual learning~\cite{Ye2020Dual, DualCos}, reinforcement learning~\cite{Yao2018Improving}, et al. Although earlier studies achieved considerable performance improvement, the relationship between method names and code summaries, which could improve code summarization significantly as we demonstrated in this work, has not been well explored yet.

As shown in Table \ref{tab:method_name_code_summary}, although the name and code summary of a method have a large difference in length,  they both can be considered as the functional description of the method code. We analyzed the method name and code summary in the dataset built by Leclair et al.~\cite{Leclair2019recommendations}, and found that averagely 50.6\% of the words in method names appear in the corresponding summaries, and 21.3\% of the words in summaries appear in the corresponding method names. For about 20\% of the methods, all the words in the method names appear in the corresponding summaries. In fact, programmers are also do some kinds of code summarization work in the process of naming a method, and they usually write code summaries based on the corresponding method name. Therefore, the task of Method Name Generation (MNG), which predicts a method name with its given body~\cite{Allamanis2016A, Alon2018code2seq, Alon2018code2vec},  can be used as an auxiliary task of code summarization to provide a useful inductive bias. Adding this auxiliary task will make the model more focused on those hypotheses that can explain both code summarization and MNG at the same time, and consequently have a potential to improve the generalization and performance of the model.

\begin{table}[tb]
  \caption{Examples of method name and code summary.}
  \centering
  \label{tab:method_name_code_summary}
  \begin{tabular}{cc}
    \hline
    Method Name & Code Summary\\
    \hline
    update state & called when a command update its state\\
    publish & publish a tt log record tt\\
    tool enabled & invoked if a tool was enabled\\
    get type & get the session manager event type\\
    \hline
\end{tabular}
\end{table}

However, poorly-defined method names, which are common in practical software projects ~\cite{host2009debugging, liu2019name}, could lead to inaccurate code summaries. As shown in Figure \ref{fig:cos_with_diff_name}, method names which contain customized abbreviations, nonstandard conjunction, or spelling errors may mislead the code summarization model. In other words, the state-of-the-art models are highly dependent on high-quality method names. In this case, if MNG could first produce a more meaningful method name, the summarization model can make further extension and polishing based on a more informative compressed draft (i.e., the generated method name). 
Inspired by the high-level concept of ``deliberation" proposed 
by Xia et al.~\cite{Xia2017Deliberation}, we incorporate another auxiliary task of method name informativeness prediction (MNIP) into a novel two-pass deliberation process in our MTL architecture, aiming to generate consistent code summaries more robustly, especially when method names are non-informative.


Last but not least, since the method name is always a part of the method code, the MNG task has a natural large-scale self-labeled training corpus, which resembles the language model in natural language processing. Language models predict the next (or masked) word in a given text sequence, which can be trained and used to improve almost all other NLP tasks, thanks to extremely large self-supervised corpora~\cite{Devlin2019Bert, Yang2019XLNet, Liu2019RoBERTa}. In a similar way, we show that 
by pre-training MNG, the self-supervised task, with an extra large-scale corpus, we could further improve code summarization.


In short, by introducing tasks MNG and MNIP, we design a deliberation multi-task learning approach, which better characterizes the relationship between method summaries and method names, to improve code summarization. To evaluate our approach, we carried out large-scale experiments on two existing datasets for Java and Python. We also collected a  huge self-labeled dataset (30.4  million pairs of method bodies and their names) from Github for evaluating the pre-training effect of task MNG. From the experiment results, we found that exploiting the inner connection between code summaries and method names can significantly improve the performance of code summarization. We believe the deliberation MTL architecture has the potential to be extended to more similar scenarios, e.g., mining inner connections of summaries and headlines of internet news.

Our main contributions are:

\begin{itemize}
\item We proposed a Deliberation Multi-task learning Approach for COde Summarization (DMACOS). To the best of our knowledge, we are the first to explore the inner connection between method names and code summaries systematically in the research of code summarization. 

\item We carried out an experiment on two large existing datasets for code summarization (on Java and Python). The experiment results show that our MTL approach can not only achieve competitive performance, but also be easily applied to some existing encoder-decoder models of code summarization and improve their performance. 

\item We studied DMACOS's performance on methods with non-informative names, and DMACOS showed significant superiority brought by its two-pass deliberation mechanism.

\item We showed extra knowledge introduced by MNG model pre-trained with large-scale self-supervised corpus can further improve the performance of the model, which can be observed even across programming languages.

\end{itemize}

\section{Proposed Approach}

\subsection{Overall Architecture}

\begin{figure}[ht]
  \centering
  \includegraphics[width=\linewidth]{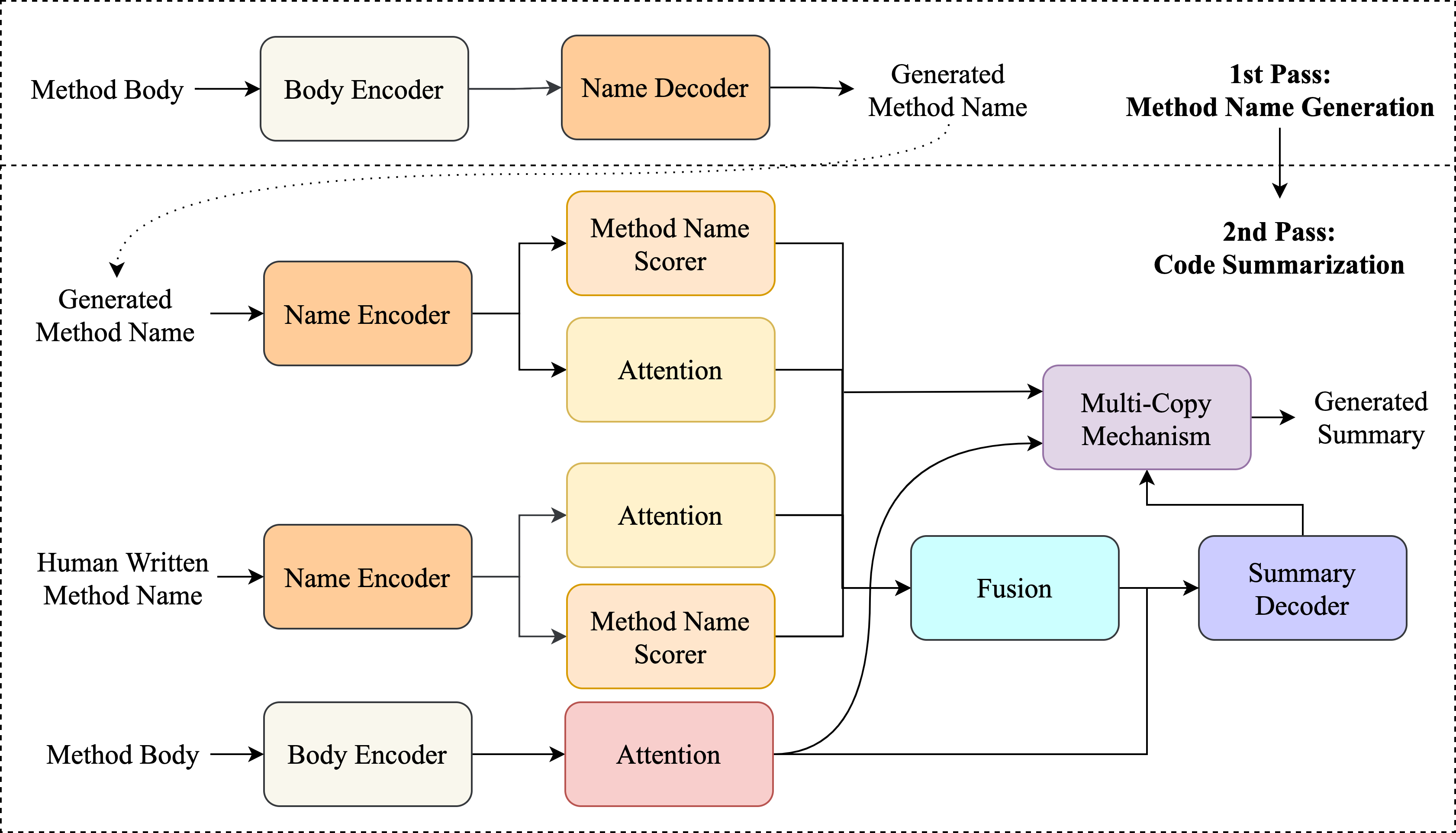}
  \caption{The overall architecture of DMACOS. In the first pass, we generate the method name (a compressed version of summary). Then the 2nd pass incorporates a deliberation process to generate the final code summary, by feeding and fusing the output of the first-pass into the summary decoder. The links from method body encoder to informativeness predictor on the second pass are omitted for readability.}
  \label{fig:framework}
\end{figure}

In this section, we first present an architecture overview of our deliberation multi-task learning approach. 
The three tasks supported by our architecture are as follows:

\begin{itemize}
\item \textbf{The task of code summarization (COS)}, which generates a natural language description for the given method, is the target task in our approach.
\item \textbf{The task of method name generation (MNG)}, which predicts a method name with the given the method source code, is the auxiliary task for code summarization. 
\item \textbf{The task of method name informativeness prediction (MNIP)}, which predicts a real number between 0 and 1 indicating how informative the method name is. We use the proportion of words in a method name appearing in the corresponding code summary as the golden informativeness score.
\end{itemize}

Like most tasks of text generation, COS and MNG can be both generally formulated as encoder-decoder models. We have a method name encoder and a method body encoder in DMACOS. The three tasks share the method body encoder, and COS and MNIP share the method name encoder. Ihe input of the method body encoder is method code with the method name masked by a special token ``$<$name$>$''. 

Note that human-written method names are not always informative, while a well-trained MNG model has the potential to generate better method names in many cases~\cite{liu2019name}. 
Therefore, in addition to parameter sharing among the three tasks in a classic MTL manner, we incorporate a ~\textbf{two-pass deliberation process} shown in  Figure~\ref{fig:framework}
into our MTL architecture, to refine method name representations fed into the second-pass code summarization. The mechanism has two main characteristics:

\begin{itemize}

\item The decoding processes of code summary decoder and method name decoder are sequential instead of parallel. Only when MNG complete the whole decoding process in the first pass, COS will start its own decoding process in the second pass.

\item We make the method name encoder and decoder share the same GRU instance, so that their intermediate states are in the same vector space. Then information from a method name and a generated method name can be fused and refined by weighted addition (or a gate) of context vectors generated by states of method name encoder and decoder during decoding, respectively. The informativeness scores generated by MNIP are normalized and then used as the weights of the fusion gate.

\end{itemize}


\subsection{Method Name Generation}


The MNG task plays two main roles. One is to improve generalization capabilities as a classic auxiliary task in multi-task learning, and the other is to contribute to providing more informative representations of method names. It contains two parts: the method body encoder and the method name decoder.

\begin{table*}[tb]
  \caption{Example of aSBT}
  \label{tab:aSBT}
  \centering
  \begin{tabular}{lccccccccc}
    \hline
    Source code & \multicolumn{9}{l}{storage\_client = Client()} \\
    original SBT sequence& \multicolumn{9}{l}{( Assign SimpleName\_storage\_client ( Call SimpleName\_Client ) Call ) Assign}\\ 
    aSBT token sequence & Assign & SimpleName & storage & client & Call & SimpleName & Client & Call & Assign\\
    aSBT type sequence &  0 & 2 & 3 & 5 & 0 & 2 & 6 & 1 & 1\\
    \hline
\end{tabular}
\end{table*}

\begin{table}[tb]
  \caption{aSBT type details}
  \label{tab:aSBT_details}
  \centering
  \begin{tabular}{cl}
    \hline
    aSBT type & description \\
    \hline
    0 & The beginning of AST node type. \\
    1 & The end of AST node type. \\
    2 & A single AST node type. \\
    3 & The beginning of token. \\
    4 & The middle of token. \\
    5 & The end of token. \\
  \hline
\end{tabular}
\end{table}

\subsubsection{Method Body Encoder}

The Method Body Encoder aims to convert the input parsed from method body to the corresponding vector representation. In a neural-network model of text summarization, input of the encoder is the token sequence of source text. Similarly, in code summarization task, the code pieces as input can be treated as word sequence\cite{Iyer2016Summarizing}. Researchers extract the word tokens in source code and put them into the encoder as input sequence. Although the text encoder achieved good results in code summarization task via observing meaningful words in code, the structural information of the source code, which is the determining factor of source code's behavior, can not be fully captured.

To address this issue, researchers have proposed various encoders specifically for the structured source code \cite{Hu2018Deep,Alon2018code2seq,Alon2018code2vec}. For example, He et al. \cite{Hu2018Deep} proposed a new structure-based traversal (SBT) which flattens the AST and ensures that the words in the code are associated with their AST node type at the same time, so that the SBT encoder can capture both the word semantic information and the code structure semantic information. Although SBT archived excellent results on code summarization task, SBT may have limitations on informaton loss cased by OOV(out Of vocabulary) words. As we all know, complex tokens under camel case rule are widely used in source code and these tokens would produce a large number of OOV words \cite{shikun2020keyword}. However, in SBT, we can not make use of such tokens easily. 

To tackle this problem, we propose a new advanced structure-based traversal (aSBT) method to traverse ASTs based SBT. Using aSBT, the abstract syntax tree (AST) can be converted to two flattened sequence: the aSBT token sequence and aSBT type sequence. TABLE~\ref{tab:aSBT} shows a simple example of aSBT and TABLE~\ref{tab:aSBT_details} shows all aSBT types and their corresponding descriptions. As shown in TABLE~\ref{tab:aSBT}, SBT token sequence and SBT type sequence represents the content information and structure information in original SBT sequence respectively.

Once the aSBT token sequence and aSBT type sequence are all ready, we use a embedding layer to convert them to aSBT token embedding sequence. So we can get a method body embeddings $x^{b} = \{ x^{b}_1, x^{b}_2, ..., x^{b}_{m}\}$ with $m$ tokens, where the upperscript $b$ represents method \textbf{b}ody.

Then a RNN layer is used to encode the embedding sequence into a semantic embedding sequence which encode the significant information needed in decoder. We choose the GRU as our RNN layer, the hidden state of the GRU at each time step $t$ is computed as:

\begin{equation}
z_t = \sigma(W_z[h_{t-1}, x_t])
\end{equation}

\begin{equation}
r_t = \sigma(W_r[h_{t-1}, x_t])
\end{equation}

\begin{equation}
\hat h_t = tanh(W[r_t \cdot h_{t-1}, x_t])
\end{equation}

\begin{equation}
h_t = (1 - z_t) \cdot h_{t -1} + z_t \cdot \hat h_t
\end{equation}

where $\sigma$ is the sigmoid function $\sigma(x) \in [0, 1]$, $h_{t-1}$ is the previous hidden state, $x_t$ is the current input embedding, and variables $W_z, W_r, W$ are the parameters of GRU. To maximize the use of encoder, we use the last hidden state of the previous encoder as initial state of current encoder. We feed the aSBT embeddings into method body encoder and then get the corresponding method body semantic embedding sequence $h^{b} = \{ h^{b}_1, h^{b}_2, ..., h^{b}_m \}$.

\subsubsection{Method Name Decoder}

The method name decoder aims to convert method body semantic embeddings into the code summary word sequence. We factorize the conditional in equation~\ref{eq:condition_word} into a product of word-level method name predictions:

\begin{equation}
\label{eq:condition_word}
p(y^{n}_1, y^{n}_2,...,y^{n}_{m}|h^{b}) = \prod_{j=1}^m p(y^{n}_j|y^{n}_{<j}, h^{b})
\end{equation}

where probability of each $y^{n}_t$\footnote{The upperscript $n$ represents method \textbf{n}ame.} is predicted based on all the words that are generated previously (i.e. $y^{n}_{<t}$) and method body representation $h^b$. More specifically, the probability is calculated as follows:

\begin{equation}
\begin{aligned}
p(y^{n}_t|y^{n}_{<t}, h^{b}) &= softmax(W_s \cdot tanh (W_t \cdot [s^{n}_t;c^{b}_t])) \\
s^{n}_t &= GRU(y^n_{t-1}, s^n_{t-1}) \\
c^{b}_t&=attention(s^{n}_t, h^{b})
\end{aligned}
\end{equation}

where $s^{n}_t$ denotes hidden state of decoder at time step $t$ and $c^{b}_t$ denotes the contextual information in generating word $y^{n}_t$ according to different encoder hidden states, which is computed as follows:


\begin{equation}
\begin{aligned}
\alpha^{b}_{ti} &= exp(s^{n}_t W_a  h^{b}_i) / \sum_j^{m} exp(s^{n}_t W_a  h^{b}_j)\\
c^{b}_{t} &= \sum_i^{m} {\alpha^{b}_{ti} h^{b}_i}
\end{aligned}
\end{equation}

\subsection{Code Summarization Model}


COS model aims to build the mappings bettween source code and the corresponding summaries. And to generate a more informative summary, COS model takes generated method name, human-written method name and method body as inputs and then introduces a deliberation process to fuze the information among them. 

\subsubsection{Method Name Encoder} Since generated method name and human-written method name are both the abstractions of source code, we believe they can benefit from sharing parameters to promote the capacity of capturing the gist of method name. To this end, we use a shared method name encoder to generate hidden state sequences for both generated method name and human-written method name. Similar to method body encoder, the embedding layer and GRU layer are used to project method name to its corresponding vectors, the vectors encode the semantic information on the method name. Hereafter, the human-written method name vectors $h^n = \{ h^n_1, h^n_2, ..., h^n_m \}$ and generated method name vectors $h^n = \{ h^{gn}_1, h^{gn}_2, ..., h^{gn}_m \}$ are generated, and the upperscript $gn$ represents \textbf{g}enerated method \textbf{n}ame.

\subsubsection{Multiple Attention} Multiple attention aims to capture the important clues from the representations generated by method name encoder and method body encoder. We use the standard attention mechanism as our attention method, which can be described as:

\begin{equation}
    w_{i}=\frac{\exp(f(k_i, q))}{\sum_{i=1}^m\exp(f(k_i, q))}
\end{equation}
\begin{equation}
    attention(q,k,v) =\sum_{i=1}^m w_i v_i
\end{equation}

The inputs of attention mechanism consists of queries and keys of dimension $d_k$, and values of dimension $d_v$. We compute the dot products of the query with all keys, and apply a softmax function to obtain the weights on the values. Here $f$ calculates the similarity between hidden states and we use bi-linear function $f(x,y)=xW_{bi}y$; we use the output $s^s_t$ of GRU layer in summary decoder at step $t$ as the query vector, which will be described with further details in section~\ref{sec:summary_decode}. Then we obtain the method body context vector $c^b=attention(s^s_t, h^b, h^b)$, the generated method name context vector $c^{gn}=attention(s^s_t, h^{gn}, h^{gn})$ and the human-written method name context vector $c^n=attention(s^s_t, h^n, h^n)$.

\subsubsection{Method Name Scorer} Method name scorer predict a scalar score by equation~\ref{eq:informativeness-predictor}, indicating how informative a method name is for the corresponding code summary. 

\begin{equation}
\label{eq:informativeness-predictor}
score(h) = W_p[h^b_{m};h_{m}]
\end{equation}

Therefore, we compute the informative scores of human-written method name and generated method name as follows:

\begin{equation}
w^n = score(h^n)
\end{equation}
\begin{equation}
w^{gn} = score(h^{gn})
\end{equation}

\subsubsection{Method Name Fusion} Method name fusion combines the generated method name context vector and human-written method name context vector with the informativeness score calculated by method name scorer:

\begin{equation}
c^{fn} = \hat{w}^n c^n + \hat{w}^{gn} c^{gn}
\end{equation}
where $\hat{w}^n$ and $\hat{w}^{gn}$ are normalized informativeness scores calculated by method name scorer.

\subsubsection{Summary Decoder} \label{sec:summary_decode}

Similar to the method name decoder, the code summary decoder learns a conditional probability as:

\begin{equation}
p_{cos}(y^{s}_t|y^{s}_{<t}) = softmax(W_s \cdot tanh (W_t \cdot [s^{s}_t;c^{b}_t;c^{fn}_t]))
\end{equation}
with $s^s_t = biGRU(y^s_{t-1}, s^n_{t-1})$ is the decoder variables for each time step in RNN layer, the upperscript $s$ represents \textbf{s}ummary.

As seen, the deliberation process mainly involves the operations of feeding generated method name into the second-pass and fusing $c^n$ and $c^{gn}$ into a refined representation $c^{fn}$ according to the predictions of MNIP at each decoding step.

\subsubsection{Multi-Copy} Finally, we employ a multi-copy component to copy words from source code. We propose the multi-copy component that copies a word w from all the method body, human-written method name and generated method name.

\begin{equation}
\begin{aligned}
p_{copy^b} &= \sum_{i:w_i=w} w^b_{t,i}      \\
p_{copy^n} &= \sum_{i:w_i=w} w^n_{t,i}      \\
p_{copy^{gn}} &= \sum_{i:w_i=w} w^{gn}_{t,i} 
\end{aligned}
\end{equation}

The final distribution is a weighted sum of the generation distribution and multi-copy distribution:

\begin{equation}
\gamma_t = sigmoid(W_f[s^{s}_t;c^{b}_t;c^{fn}_t] + b_f)
\end{equation}
\begin{equation}
p = \gamma_t p_{cos} + \frac{1}{3}(1-\gamma_t)(p_{copy^b} + p_{copy^n} + p_{copy^{gn}})
\end{equation}

\subsection{Pre-Training} 

Since the performance of COS task is highly dependent on the performance of MNG and MNIP tasks, we employ a pre-training process to pre-optimize the parameters of MNG and MNIP models.

In task MNG, we first extract the method name from the source code, so we can get a new training corpus of pairs between source code and method name: $S^{mng} = \{ x^{(i)}, y^{(g, i)} \}$. With the given training corpus $S^{mng}$, the COS task's training objective is to minimize the negative log-likelihood of the training data with the respect to all parameters, as denoted by $\theta$,

\begin{equation}
\begin{aligned}
  loss_{mng} = - \sum_{i=1}^{|S^{mng}|}  \sum_{j=1}^{|y^{(g, i)}|} log P(y^{(g, i)}|x^{(i)}; y^{(g, i)}_{<j}; \theta)
\end{aligned}
\end{equation}

In task MNIP, we use the proportion of words in method name also appearing in the corresponding code summary as the golden informativeness score, so we can get a new MNIP corpus: $S^{mnip} = \{ x^{(i)}, y^{(ip, i)} \}$. We minimize a mean-square error criterion over the training set:

\begin{equation}
\begin{aligned}
  loss_{mnip} = \frac{1}{|S^{mnip}|} \sum_{i=1}^{|S^{mnip}|} (y^{(ip, i)} - w^{(n, i)})^2
\end{aligned}
\end{equation}

\subsection{Multi-Task Training} 

Given a training corpus of code-summary pairs: $S^{cos} = \{ x^{(i)}, y^{(s, i)} \}$, the training details introduced as follows. With the given training corpus $S$, the COS task's training objective is to minimize the negative log-likelihood of the training data with the respect to all parameters, as denoted by $\theta$,

\begin{equation}
\begin{aligned}
  loss_{cos} = - \sum_{i=1}^{|S^{cos}|} \sum_{j=1}^{|y^{(s, i)}|} log P(y^{(s, i)}|x^{(i)}; y^{(s, i)}_{<j}; \theta)
\end{aligned}
\end{equation}

Our final joint learning objective becomes:

\begin{equation}
  loss = loss_{cos} + \alpha loss_{mng} + \beta loss_{mnip}
\end{equation}

\section{Experiments}

\subsection{Dataset Details}

To ensure the generality of our experiment results, we use three datasets to evaluate our approach. 

\subsubsection{Java Summary Dataset} This is the standard Java dataset provided by Leclair et al.~\cite{Leclair2019recommendations}, which contains over 2.1 million Java methods and associated comments. Following Leclair et al.~\cite{Leclair2019A}, we split it into training, validation, and testing sets in proportion of 90 : 5 : 5 after shuffling the pairs. 

\subsubsection{Python Summary Dataset.} This dataset is built by Barone et al.~\cite{Barone2017A}. It is a diverse parallel corpus of Python methods with their documentation strings. The dataset contains 113,108 code-comment pairs, and the pairs which can not be parsed by Python3 are removed. We use the same split of training, validation, and testing sets as the original settings in Barone et al.~\cite{Barone2017A}, consisting of 109,108 training examples, 2,000 validation examples and 2,000 test examples.

\subsubsection{Java Method Name Dataset.} This dataset is collected from popular Java projects that have at least 100 stars on GitHub by ourselves, which is only used for pre-training of the MNG task. We build this dataset to see whether pre-training of MNG task on large-scale data can improve the performance. This dataset has more than 30.4 million Java methods and corresponding method names.

\subsection{Experiment Settings}

Following Leclair et al.~\cite{Leclair2019A}, we set the dimensionality of the GRU hidden states, method body embeddings and summary embeddings to 256, 100 and 100, respectively . The method name share the same vocabulary and embedding space with the summary. 
The hyper-parameter $\alpha$ and $\beta$ are set to 0.1 and 0,1 since they achieved best performance.
The maximum lengths for method name sequences, method body sequences and summary sequences are 10, 300, 13 for Java, and 10, 100 and 20 for Python, each covering at least 90\% of the training set. Sequences that exceed the maximum will be truncated and the shorter sequences are padded with zeros. The vocabulary size of the method body, and summary are 50000, 44707 for Java and 50400, 31350 for Python according to Leclair et al.~\cite{Leclair2019A} and Yao et al.~\cite{Yao2018Improving} respectively. For each approach, we computed performance metrics for the model after each epoch against the validation set. Then we chose the model after the epoch with the highest validation performance, and computed performance metrics for this model against the testing set. Adam is used for parameter optimization. The learning rate is set to 0.001. Our model is implemented in PyTorch and trained on Tesla T4.

\subsection{Metrics} 
Following previous works, we evaluate the performance of DMACOS and baselines based on BLEU4 ~\cite{Papineni2002BLEU}, METEOR ~\cite{Banerjee2005METEOR} and ROUGE-L~\cite{lin2004rouge}, all of which are widely used in evaluating performance of text generation tasks. BLEU4 score is a popular accuracy-based measure for machine translation. It calculates the similarity between the generated sequence and reference sequence by counting the n-grams that appear in both the candidate sequences and the reference sequence. METEOR measure is the harmonic average of precision and recall, with argument that recall-based measures can be more correlative to manual judgement than accuracy-based measures like BLEU. ROUGE-L takes into account sentence level structural similarity naturally and identifies the longest co-occurring in sequence n-grams automatically.

\subsection{Research Questions}

Our motivating intuition is that leveraging method name for two-pass deliberation multi-task learning can improve the results of existing code summarization techniques. Therefore, in our experiment, the major goal is to validate this intuition and find out how much enhancement our technique can bring to various existing techniques. So we try to answer the following research questions to achieve this goal. 

\subsubsection{RQ1: How effective is DMACOS compared to state-of-the-art neural-network-based models?}

There are many neural-network-based models for code summarization, and we selected representative ones as baselines and will be described in Section~\ref{subsec:baseline}. We evaluated the performance of DMACOS in Java summary dataset and Python summary dataset using metrics described earlier.

\subsubsection{RQ2: Can DMACOS improves performance of existed encoder-enhaning models?}

Researchers in the code summarization field have proposed a variety of neural models, in which many of them tried to design better code encoder to capture more accurate semantics of code. These encoder-enhaning models can be easily equipped with our MTL architecture. To demonstrate the effectiveness of DMACOS in terms of improving encoder-enhaning models, we implemented their DMACOS-equipped versions, and compared the evaluation metric scores with those of their original versions based on the aforementioned metrics.

\subsubsection{RQ3: How does MTL, deliberation and MNIP affect the performance of DMACOS?}

To demonstrate how does the MTL, deliberation and MNIP affect the performance of DMACOS, we removed multi-task learning, two-pass deliberation process and MNIP task from the DMACOS and implement three variant models. We compared these three models based on the aforementioned metrics.

\subsubsection{RQ4: How does DMACOS perform in the scenarios of poorly-named methods?}

Since poorly-defined method names are common in practical software projects, we designed \textbf{RQ4} to study how a model performs if the method names are poorly-defined. To simulate the scenario of handling non-informative method names, we replaced method names with a universal special ``UNK'' token on source code and compared the performance of DMACOS and Dual model on such replaced dataset.

\subsubsection{RQ5: Can DMACOS improves performance of code summarization further if the MNP pre-training is performed on a larger-scale training dataset?}

One major advantage of DMACOS is that it can take advantage of the naturally available large-scale dataset for the MNP task. When answering the first two research questions, we show that DMACOS can already enhance performance of code summarization by pre-training MNP task on the same dataset. \textbf{RQ5} tries to investigate whether a larger pre-training dataset for the MNP task can further enhance experiment results. To answer \textbf{RQ5}, we first pre-trained a Seq2Seq model which consist of SBT encoder, text encoder and method name decoder on the Java method name dataset for MNP task. Then we initialize the parameters in SBT encoder and text encoder with the pretrained parameters in baseline model and our model. Finally we fine-tuning these models on the standard Java summary dataset in different dataset size. After that, We compare these models based on the aforementioned metrics.

\subsubsection{RQ6: How does MNG pre-training perform in the scenarios of cross-programming-language training?}

For less popular programming languages, it is often difficult to construct a large training dataset for either MNP or code summarization tasks. Therefore, we designed \textbf{RQ6} to study how a model performs if it is pre-trained (for the MNP task) on one programming language (which is more popular), and tuned (for the code summarization task) on another language. To answer this question, we trained a DMACOS-based model on the python dataset with MNP pre-training on the Java dataset. 

\subsection{Baselines} \label{subsec:baseline}

We compare our approach with the following and state-of-the-art models as baselines of our evaluation.


\begin{itemize}

\item \textbf{CodeNN} CodeNN\cite{Iyer2016Summarizing} is an end-to-end code summarization approach. They use LSTM to generate summaries given code snippets. At each time step, CodeNN generates a word by applying the attention mechanism, which computes a weighted sum of the code token embeddings.

\item \textbf{DeepCom} DeepCom\cite{Hu2018Deep} is a model based on the popular attention-based seq2seq NMT systems. The model takes the structure-based traversal (SBT) representation as input, which converts the abstract syntax tree (AST) to a flattened sequence.

\item \textbf{Ast-attendgru} Ast-attendgru\cite{Leclair2019A} is an approach which uses both original code text and SBT representation as input. The approach builds a GRU-based-encoder-decoder model with two encoders and one decoder. At each time step of decoding, attention mechanism is applied between the outputs of all the encoders and the decoder.

\item \textbf{DRL} DRL\cite{Yao2018Improving} incorporate an abstract syntax tree structure as well as sequential content of code snippets into a deep reinforcement learning framework. And the BLEU metric score is used as the advantage reward to provide global guidance for explorations.

\item \textbf{Dual Model} Dual model~\cite{DualCos} improve code summarization by introducing a dual code generation task and designing a regularization terms based on the dualities of code summarization and code generation on the probability and attention weights.

\end{itemize}

\section{Experiment Results}

\begin{table*}[tb]
  \caption{Experimental Results on the Java and Python datasets. Relative improvements we concerned are all applied T-test and all p-values are smaller than 0.01, indicating significant increases.}
  \label{tab:performance_all}
  \centering
  \begin{tabular}{lrrrrrr}
    \hline
    \multirow{2}*{Approaches}&\multicolumn{3}{c}{\textbf{Java}}&\multicolumn{3}{c}{\textbf{Python}} \\
    \cline{2-7}
    & \small{BLEU4}  & \small{METEOR} & \small{ROUGE-L} & \small{BLEU4} & \small{METEOR} & \small{ROUGE-L} \\
    \hline
    CodeNN               & 6.85 & 13.9 & 32.1 & 6.84 & 13.1 & 28.3 \\
    DeepCom              & 7.70 & 15.2 & 35.0 & 7.90 & 13.4 & 29.9 \\
    Ast-attendgru        & 11.3 & 17.9 & 39.5 & 8.71 & 16.5 & 34.8 \\
    DRL  & 12.1 & 18.6 & 40.7 & 11.6 & 16.3 & 35.6 \\
    Dual Model           & 12.3 & 19.2 & 41.2 & 11.9 & 17.5 & 36.8 \\
    DMACOS               & \textbf{12.9}& \textbf{20.1} & \textbf{42.3}
                         & \textbf{13.2}& \textbf{19.4} & \textbf{38.5}\\
    \hline

    DMACOS(DeepCom)          & 9.81 & 18.0 & 39.4 & 10.4 & 17.1 & 35.7 \\
    DMACOS(Ast-attendgru)& 12.4 & 20.2 & 44.4 & 12.6 & 17.6 & 36.5 \\
    \hline
    DMACOS w/o MTL       & 11.1 & 18.6 & 38.9 & 8.70 & 16.1 & 33.2 \\
    DMACOS w/o two-pass  & 12.2 & 19.1 & 38.8 & 12.1 & 18.1 & 37.9 \\
    DMACOS w/o MNIP      & 12.5 & 19.7 & 40.8 & 12.7 & 18.9 & 38.3 \\
    \hline
    pre-trained DMACOS & \textbf{13.7} & \textbf{21.0} & \textbf{45.9} & \textbf{14.1} & \textbf{19.7} & \textbf{39.5}\\  
    \hline
\end{tabular}
\end{table*}

\begin{table*}[tb]
  \caption{Experimental results on the Java and Python dataset with method name masked.}
  \label{tab:performance_masked}
  \centering
  \begin{tabular}{lrrrrrr}
    \hline
    \multirow{2}*{Approaches}&\multicolumn{3}{c}{\textbf{Java}}&\multicolumn{3}{c}{\textbf{Python}} \\
    \cline{2-7}
    & \small{BLEU4}  & \small{METEOR} & \small{ROUGE-L} & \small{BLEU4} & \small{METEOR} & \small{ROUGE-L} \\
    \hline
    Dual                        & 6.34 & 15.1 & 32.7 & 4.78 & 12.1 & 25.1 \\
    DMACOS w/o MTL              & 5.75 & 12.4 & 29.6 & 4.87 & 11.6 & 25.2 \\
    DMACOS w/o two-pass         & 4.16 & 10.1 & 24.7 & 5.92 & 10.4 & 24.7 \\
    DMACOS w/o MNIP             & 7.39 & 16.1 & 33.3 & 7.11 & 11.9 & 27.5 \\
    DMACOS                      & \textbf{8.21} & \textbf{17.6} & \textbf{37.0} 
                                & \textbf{8.25} & \textbf{13.3} & \textbf{29.6} \\
    \hline
\end{tabular}
\end{table*}

\subsection{Overall Results}

Our first research question is: \textit{How effective is MACOS compared to state-of-the-art neural-network-based models?} To answer \textbf{RQ1}, We evaluated and compared MACOS against the baselines CodeNN, DeepCom, Ast-attendgru, DRL and Dual model, among which the Dual model is state-of-the-art code summarization method. TABLE~\ref{tab:performance_all} reports the performance of the baselines and our model, and our model outperforms all baselines in the three evaluation metrics. 

As shown in TABLE~\ref{tab:performance_all}, the BLEU4, METEOR and ROUGE\_L scores of DMACOS are higher than all baselines in both Java and Python datasets. Compared with a reinforcement learning  model (DRL), the relative enhancements for BLEU4, METEOR and ROUGE\_L on Java and Python dataset are 6.6\%, 8.0\%, 3.9\%, 13.8\%, 16.6\% and 8.1\%, respectively. Compared with a dual learning model (Dual Model), the relative enhancements for BLEU4, METEOR and ROUGE\_L on Java and Python dataset are 4.8\%, 4.7\%, 3.2\%, 11.0\%, 8.6\% and 4.6\%, respectively. The results verify that code summarization can be improved significantly by better exploiting the relationship between method code summaries and method names.

\subsection{Applying DMACOS to other models}

Our second research question is: \textit{Can DMACOS improves performance of existed encoder-enhaning models?} To answer \textbf{RQ2}, we evaluated and compared DeepCom and Ast-attendgru with their MACOS-equipped versions on the aforementioned metrics. In particular, we plug their encoder into DMACOS by simply replacing the original method body encoder, so that we obtain DMACOS(DeepCom) and DMACOS(Ast-attendgru).

The seventh and eighth rows of results in TABLE~\ref{tab:performance_all} show that DMACOS led to significant enhancement on DeepCom and Ast-attendgru, e.g., with the relative increment on BLEU4 in Java and Python datasets as 27.4\%, 31.6\%, 9.7\% and 44.6\% respectively. It is notable that applying DMACOS method to Ast-attendgru even achieves a better performance than Dual Model, the latest competitive approach.

\subsection{Ablation Study}


Our third research question is: \textit{How does MTL, deliberation and MNIP affect the performance of DMACOS?} To answer \textbf{RQ3}, We obtain three model variants with a subset of DMACOS's key components: (1) \textbf{DMACOS w/o MT}L with the entire MTL design removed, (2) \textbf{DMACOS w/o two-pass} with only MNG  as an auxiliary task in a classic single-pass MTL architecture, and (3) \textbf{DMACOS w/o MNIP} that employ the two-pass mechanism that feeding generated method name information of the first pass directly into the summary decoder without the refining process guided by MNIP. The performance of the three DMACOS variants is also reported in TABLE~\ref{tab:performance_all}, from which we have the following observations:

\begin{enumerate}

  \item It is as expected that removing all MTL components makes the performance degrade significantly (e.g., 11.1 of DMACOS w/o MTL v.s. 12.9 of DMACOS in terms of BLEU score on Java dataset).  

  \item Compared with DMACOS w/o two-pass, DMACOS w/o MNIP gain a relative improvement for BLEU4, METEOR, and ROUGE\_L of 2.5\%, 3.1\%, 5.2\% (Java dataset) and 5.0\%, 4.4\%, and 1.1\%  (Python dataset), verifying the effectiveness of our two-pass design even with only task MNG. 

  \item Compared with DMACOS w/o MNIP, full DMACOS achieves further improvement (e.g., a relative BLEU4 improvement of 3.2\% and 3.9\% on Java and Python dataset respectively) via the MNIP-guided refinement on method name representations. This ablation study clearly demonstrates the effectiveness of all designs involving introducing and utilizing MNG and MNIP.

\end{enumerate}

\subsection{Impact of method name informativeness}

\begin{figure*}[!t]  
\centering
\includegraphics[width=\linewidth]{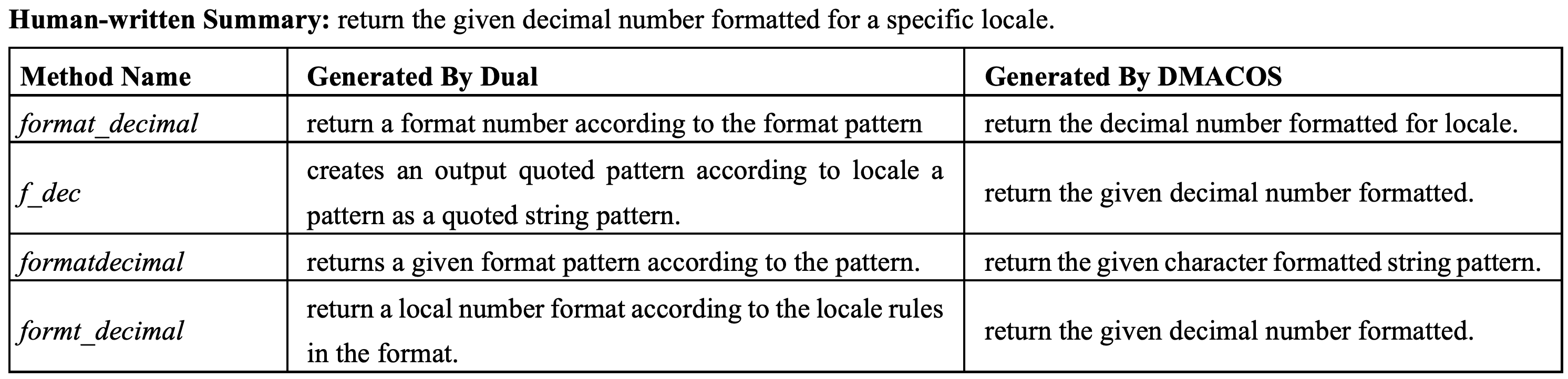}
\caption{Summaries generated by Dual Model and DMACOS with different method names. The corresponding source code is shown in Figure~\ref{fig:cos_with_diff_name}.}
\label{fig:cos_with_diff_name2}
\end{figure*}

Our fourth research question is: \textit{How does DMACOS perform in the scenarios of poorly-named methods?} To answer \textbf{RQ4}, we replaced method names with a universal special ``UNK'' token in the testing dataset to simulate the scenario of handling non-informative method names. Under this experiment setting, the performance of DMACOS, its variants, and Dual Model are shown in Table ~\ref{tab:performance_masked}, from which we have three observations. 

\begin{enumerate}

  \item Significant performance degradation occurs in all models, indicating the significance of method names in code summarization. 

  \item Interestingly, the model using a classic MTL architecture without the two-pass mechanism (DMACOS w/o two-pass) performs worse than the model without MTL architecture (DMACOS w/o MTL), e.g., on the Java dataset.  This observation reflects the side effects of being too dependent on well-defined method names. 

  \item The degradation of DMACOS is much slighter than that of DMACOS w/o two-pass (e.g., about one-third v.s. two-third of relative decrement in terms of BLEU score on the Java dataset), showing that DMACOS can alleviate the impact of non-informative method names better. 

  \item Comparison results among models (e.g., DMACOS v.s. DMACOS w/o MNIP, DMACOS v.s. Dual Model) generally follow a similar pattern with the main results on standard experiments.

\end{enumerate}

Note that to evaluate the impact of method name informativeness on DMACOS, another straightforward way is dividing samples in the test set into some subsets based on their informativeness. However, we find the quality of code summaries in each subset diversify enormously, making it difficult to make fair comparisons among results on these subsets, though we find that DMACOS always outperforms the Dual model on each subset.

As a further qualitative analysis, recapping the case illustrated in Figure~\ref{fig:cos_with_diff_name} in Section~\ref{Introduction}, here we feed those non-informative method names into DMACOS. Summaries generated by our DMACOS and Dual Model are shown in Figure~\ref{fig:cos_with_diff_name2}, and DMACOS's output is obviously more consistent with human-written summaries.

\subsection{Pre-Training on a Larger Dataset}

Our fifth research question is: \textit{Can DMACOS improves performance of code summarization further if the MNP pre-training is performed on a larger-scale training dataset?} To answer \textbf{RQ5}, we first performed pre-training for the MNP task on the large dataset we collected from Github. Then we performed the multi-task learning on the standard dataset of Java language. This technique is also called fine-tuning in deep learning community.

The results of the model with MNG pre-trained (pre-trained DMACOS) are shown in the part of Java dataset on last row in TABLE~\ref{tab:performance_all}. The results show that on all datasets and all metrics, pre-trained DMACOS performs better than DMACOS (e.g., 13.7 v.s. 12.9 in terms of BLEU score on the Java summary dataset).  The results also show that MNP large-scale corpus has a large tolerance to noise, allowing us to utilize the large-scale corpus without carefully designing filtering rules, which is an advantage brought by self-labeled dataset.

\subsection{Cross-Programming-Language Training}

Our sixth research question is: \textit{How does MNG pre-training perform in the scenarios of cross-programming-language training?} To answer \textbf{RQ6}, we trained a MACOS-based model on the python dataset with pre-trained MNP from corpus of Java. The results of the model with MNG pre-trained (pre-trained DMACOS) are shown in the part of Python dataset on last row in TABLE~\ref{tab:performance_all}. The results indicate that knowledge captured by the MNG task can be utilized even across programming languages. The reason is that the text and structure semantics within the code do have similarities among different programming languages and the MNG task can characterize them well via large-scale corpus to improves performance. 

\subsection{Threats to Validity}

One threat to the construction validity of our experiment is that we use automatic metrics to evaluate the quality of summaries. All of these metrics may have limitations on simulating the quality of generated code summary. However, since human-based evaluation is not scalable and can still be subjective, these metrics are widely used in prior research efforts on text generation (including code-summary generation). To reduce this threat, we used multiple metrics in our evaluation to avoid limitations of a specific metric affecting our results. To further reduce this threat, we plan to perform human-based evaluation in the future and combine human-based evaluation metrics with automatic evaluation metrics. 

The major threat to internal validity is that we use the first sentence of code documentation as our ground truth. This is also a common practice in prior research efforts on code summarization\cite{Leclair2019A}. This approach will inevitably introduces noise into the data, e.g., mismatches between methods and the one-sentence summarization. It is necessary to investigate how to build a better parallel corpus. And we believe that with the in-depth application of neural network in code summarization, we will be able to build models with excellent performance on much fewer parallel corpora, and MACOS is a solid step toward this direction. 
Another threat to internal validity is that we did not perform cross-validation. We attempt to mitigate this risk by using random samples to split the training/validation/testing sets, a different split could result in different performance. This risk is common among summarization experiments due to very high training computation costs.

The major threat to external validity is that our experiment results may be applicable only to the dataset we used in our experiment. To reduce this threat, we used two very large evaluation dataset in two different popular programming langauges: Java and Python. To further reduce this threat, in the future, we plan to perform more experiments on more datasets in other programming languages. 

\section{Related Work}

\subsection{Code Summarization}

Automatic code summarization now is an important and rapidly-growing research topic in the community of software engineering and natural language processing. For traditional techniques,  we direct refers to a comprehensive survey by Nazar et al. ~\cite{DBLP:journals/jcst/NazarHJ16}. Since the first neural model of code summarization was proposed by Iyer et al. \cite{Iyer2016Summarizing}, we have witnessed the introduction of the latest neural technologies for text generation tasks (e.g., machine translation and text summarization) into this research field recently. For example, Wei et al.~\cite{DualCos} and Ye et al. ~\cite{Ye2020Dual} introduced the code generation task to improve code summarization task via dual learning ~\cite{He2016Dual}. Hu et al. ~\cite{Hu2018Summarizing} proposed an API sequence encoder to assist code summarization by transferring learning of API knowledge. Yao et al. ~\cite{Yao2018Improving} adopted Tree-RNN and reinforcement learning to enhance BLEU score of code summaries. Chen et al. ~\cite{Chen2018VAE} designed a neural framework for code summarization and code retrieval based on Variational Auto Encoders (VAEs).

\subsection{Method Name Prediction}

Method name prediction problem can be seen as structured prediction problem. Given method codes, a sequence of method name is needed to be predicted. Regarding both method code and method name as a sequence, the seq2seq paradigm\cite{Sutskever2014Sequence} can be used to model the problem. Allamanis et al.\cite{allamanis2015suggesting} use a logbilinear context model to generate method names. They define a local context capturing tokens around the current token, and a global context which is composed of features from the method code. As mentioned above, Allamanis et al.\cite{Allamanis2016A} propose an attentional neural network that employs convolution on the input tokens to detect features for method name prediction.The abstract syntax tree (AST) of method code contains meaningful structural properties, thus models based on tree structures like Tree-LSTM\cite{Kai2015Improved} can encode the method code in the form of AST. To better encode source code, Alon et al. \cite{Alon2018code2seq} present code2seq model, which leverages the compositional paths in AST and uses attention to select the relevant paths while decoding. In a subsequent paper\cite{Alon2018code2vec}, the path representation is computed using Long Short Term Memory network (LSTM) instead of a single layer neural network.

\subsection{Code Representation}

Code representation and code summarization are two closely related studies. The code representation studies how to generate word vectors for each token or an overall representation for a piece of code. Harer et al.\cite{Harer2018Automated} use word2vec to generate word embedding for C/C++ tokens for software vulnerability prediction.The token embedding is used to initialize a TextCNN model for classification. Xie et al. \cite{Xie2019DeepLink} and Zhang et al. \cite{zhang2020exploiting} build a code knowledge graph and learn code context representation based on this knowledge graph. The learned representations are used to recover the missing links between issues and commits. Mou et al.\cite{mou2016convolutional} learn distributed vector representations using custom convolutional neural networks to represent features of snippets of code, then they assume that student solutions to various coursework problems have been intermixed and seek to recover the solution-to-problem mapping via classification. Li et al.\cite{Li2015Gated} learn distributed vector representations for the nodes of a memory heap and use the learned representations to synthesize candidate formal specifications for the code that produces the heap. Alon et al.\cite{Alon2018code2vec} compute Java method embeddings by decomposing code to a collection of paths between two leaf nodes in its abstract syntax tree, and learning the atomic representation of each path simultaneously with learning how to aggregate a set of them. The method embedding is used to predicting method names.

\subsection{Multi-Task Learning}

MTL is heavily used in machine learning and natural language processing tasks. MTL learning aims to help improving the learning of a model by leveraging the domain-specific knowledge contained in the training signals of related tasks\cite{caruana1997multitask}. Usually, relatedness among tasks are learned in two ways in deep neural networks: hard parameter sharing and soft parameter sharing of hidden layer\cite{ruder2017overview}. Hard parameter sharing MTL was first proposed in \cite{Caruana1993Multitask}, which shares the hidden layer between all tasks and keep task-specific output layers. Collobert et al.\cite{collobert2008unified} describe a single convolutional neural network architecture trained jointly on NLP tasks such as part-of-speech tags, chunks, named entity tags, and semantic roles. Zheng et al.\cite{zheng2018same} propose a module in which all tasks share the same sentence representation and each task can select the task-specific information from the shared sentence representation with attention mechanism. On the other hand, each task in soft parameter MTL contains its own model and parameters, and the parameters are encouraged to be similar with regularization. Misra et al.\cite{misra2016cross} connect two separate networks in a soft parameters sharing way. Then the model leverages a unit called cross-stitch to determine how to combine the knowledge learned in other related tasks to task-specific networks.

We first introduced MTL into the study of code summarization. Hu et al.\cite{Hu2018Summarizing} use a fine-tuning approach; however, their pre-training is mainly used to capture features (e.g., API call sequences) from the input data of the same task, rather than share knowledge of different tasks. They declared their work as a transfer learning approach, but we believe it is more about designing better code features, since datasets used for pre-training and fine-tuning are generally the same. We believe that our research provides a reference for code learning tasks based on deep learning. The method of mining the association between different tasks and large-scale corpus pre-training has great potential in the field of program representation and comprehension.

\section{Conclusions and future work}


In this paper, we leverage method name generation (MNG) and method name informativeness prediction (MNIP) to improve code summarization via a two-pass deliberation multi-task learning approach. In our approach DMACOS, MNG not only serves as a highly-related auxiliary task that provides beneficial inductive bias, but also helps yielding better representations of method names with the useful guidance from task MNIP in a two-pass way. Empirical results show DMACOS could improve model performance and alleviate the impact of non-informative method names.

The novelty of DMACOS mainly lies in the idea of introduction and effective utilization of task MNG and MNIP.  It is easy to be applied to other code summarization models as demonstrated in the experiments as well as other potential scenarios of text generation. Taking summarization on internet news as an example, we can introduce generation and informativeness prediction of news headlines as auxiliary tasks in the same manner as DMACOS, of which the headline generation is also a self-supervised task. More sophisticated deliberation designs can so be incorporated into our architecture for these scenarios, which is our future work.

\section{Acknowledgement}

This research was supported by the Ministry of Industry and Information Technology of the People's Republic China (No. TC190H46G/1). Any opinions, findings, and conclusions expressed herein are the authors and do not necessarily reflect those of the sponsors.

\bibliographystyle{IEEEtran}
\bibliography{references}

\end{document}